\documentclass[10pt,twocolumn,letterpaper]{article}

\pdfoutput=1

\usepackage{cvpr}
\usepackage{times}
\usepackage{epsfig}
\usepackage{graphicx}
\usepackage{amsmath}
\usepackage{amssymb}

\usepackage{multirow}
\usepackage{booktabs}

\usepackage[pagebackref=true,breaklinks=true,letterpaper=true,colorlinks,bookmarks=false]{hyperref}

\cvprfinalcopy % *** Uncomment this line for the final submission

\def\cvprPaperID{14} % *** Enter the CVPR Paper ID here

\ifcvprfinal\fi
\begin{document}

%%%%%%%%% TITLE
\title{Mimicry: Towards the Reproducibility of GAN Research}

\author{Kwot Sin Lee \quad\quad Christopher Town\\
University of Cambridge\\
Department of Computer Science and Technology\\
{\tt\small \{ksl36, cpt23\}@cam.ac.uk}
% \and
}

\maketitle
%\thispagestyle{empty}

%%%%%%%%% ABSTRACT
\begin{abstract}
    Advancing the state of Generative Adversarial Networks (GANs) research requires one to make careful and accurate comparisons with existing works. Yet, this is often difficult to achieve in practice when models are often implemented differently using varying frameworks, and evaluated using different procedures even when the same metric is used. To mitigate these issues, we introduce Mimicry, a lightweight PyTorch library that provides implementations of popular state-of-the-art GANs and evaluation metrics to closely reproduce reported scores in the literature. We provide comprehensive baseline performances of different GANs on seven widely-used datasets by training these GANs under the same conditions, and evaluating them across three popular GAN metrics using the same procedures. The library can be found at \href{https://github.com/kwotsin/mimicry}{https://github.com/kwotsin/mimicry}.
\end{abstract}

\vspace{-0.2cm}
\section{Introduction}
\paragraph{Motivation.} Extraordinary progress has been made in the field of generative modelling following the introduction of Generative Adversarial Networks (GANs) \cite{goodfellow2014generative}, a form of generative model known for its sample efficiency in generating high resolution data \cite{odena2019open}. In particular, GANs are known for their creative applications in areas such as image synthesis \cite{shaham2019singan, karras2019style, park2019gaugan, karras2019analyzing} image-to-image translation \cite{zhu2017unpaired, zhu2017toward, lee2018diverse, lee2020drit++}, video-to-video synthesis \cite{wang2018video, wang2019few, chan2019everybody}, and even 3D view synthesis \cite{nguyen2019hologan, noguchi2019rgbd, wiles2019synsin}. Apart from the unique insights into these problems, the success of such applications often relies on fundamental progress in GAN research, such as finding novel ways to improve GAN training stability. This naturally begs the question of how one measures progress in GAN research, especially when it is difficult to compare across GANs implemented using different frameworks, trained under different conditions, and evaluated using different methods even when the same quantitative metric is used. In this work, we aim to mitigate these issues by providing:
\begin{enumerate}
    \item A lightweight and extensible library implementing common functions for training and evaluating GANs, so researchers can focus on GAN implementation and avoid re-implementing boilerplate code.
    We support popular GAN metrics such as Inception Score (IS) \cite{salimans2016improved}, Fréchet Inception Distance (FID) \cite{heusel2017gans}, and Kernel Inception Distance (KID) \cite{binkowski2018demystifying}.
    \item Standardized implementations of popular GANs using a single code base and framework, with a focus on reproducing scores reported in the literature. This allows researchers to build on top of these GANs easily without running code from different code bases.
    \item Baseline performances of popular GANs of the same model size, trained under the exact same conditions, and evaluated using the same method across multiple metrics and seven datasets commonly used in GAN research. We aim to allow for easy comparison of scores as transparently as possible.
\end{enumerate}
\section{Overview}
\subsection{Approach}
\paragraph{Models.}
We provide standardized implementations of different GANs: Deep Convolutional GAN (DCGAN) \cite{radford2015unsupervised}, Wasserstein GAN with Gradient Penalty (WGAN-GP) \cite{gulrajani2017improved}, Spectral Normalization GAN (SNGAN) \cite{miyato2018spectral}, Conditional GAN with Projection Discriminator (cGAN-PD) \cite{miyato2018cgans}, Self-supervised GAN (SSGAN) \cite{chen2018self}, and InfoMax-GAN \cite{lee2019infomax}. The first four GANs are standard GANs often implemented in practice as baselines for experiments, whereas the latter two are recent state-of-the-art unconditional GANs, of which we plan to include more in future. To verify the  correctness of our implementations, we first re-implement these GANs to ensure they reproduce reported scores, before producing the baseline results for benchmarking. 

\paragraph{Metrics.}
We provide three different metrics for comparisons: Inception Score (IS), Fréchet Inception Distance (FID), and Kernel Inception Distance (KID). IS is known to correlate well with human assessment for the quality of images \cite{salimans2016improved}, since high probabilities produced by an ImageNet \cite{deng2009imagenet} pre-trained Inception \cite{szegedy2016rethinking} network - a proxy for image realism - would result in a high value of IS. On the other hand, FID aims to measure the diversity of generated images by measuring the Wasserstein-2 distance between the Inception features, assuming their distributions take the form of a multivariate Gaussian. However, FID estimates are highly biased \cite{binkowski2018demystifying}, since using a small sample size can severely underestimate the true FID value. This is problematic when one could obtain spurious improvements from simply using a larger sample size, making it difficult to compare against prior works. A large-scale study done in \cite{kurach2018large} further highlighted how reported FID scores are often mismatched this way.  To mitigate this, we additionally assess GANs using KID, which measures the squared Maximum Mean Discrepancy (MMD) \cite{gretton2012kernel} between the Inception features using a polynomial kernel. KID is strongly correlated with FID but produces unbiased estimates \cite{binkowski2018demystifying}, which makes it fairer for comparing different models.

\paragraph{Datasets.}
We experiment on seven different datasets commonly used in GAN research: CIFAR-10 \cite{krizhevsky2009learning}, CIFAR-100 \cite{krizhevsky2009learning}, STL-10 \cite{coates2011analysis}, CelebA \cite{liu2015faceattributes} at two different resolutions, ImageNet \cite{deng2009imagenet}, and LSUN-Bedroom \cite{yu15lsun}. 

\subsection{Library Design}
\paragraph{Object-oriented design.} We implement all components in the training pipeline in an object-oriented manner. All models are subclasses of \texttt{BaseModel}, which has basic functions like saving and restoring of checkpoints. A \texttt{BaseGenerator} and \texttt{BaseDiscriminator} then inherit from \texttt{BaseModel} and implement a basic \texttt{train\_step} function for one update step (which can be overriden). These base GAN classes have an abstract method \texttt{forward} that needs to be overriden by subclasses. For each GAN variant, it has a base class inheriting from \texttt{BaseGenerator} or \texttt{BaseDiscriminator}, implementing specific functions such as gradient-penalty loss computation. To implement a GAN variant for different datasets, one would inherit from this base class to implement only the layers and forwarding method. Doing so, we maximize code-reuse when implementing the same GAN for different datasets. A \texttt{Trainer} object then assembles various parts of the pipeline, such as the models and optimizers, to construct the training pipeline. A tutorial for usage can be found at \href{https://mimicry.readthedocs.io/en/latest/guides/tutorial.html}{https://mimicry.readthedocs.io/en/latest/guides/tutorial.html}.

\paragraph{Training.} We support the training of all seven datasets with variable sizes by loading them with a single function, using common preprocessing methods for training GANs. Furthermore, while it is common to measure training iterations in terms of the number of epochs, most GAN papers report scores according to the number of training steps taken by the generator instead. Thus, the backend of the \texttt{Trainer} object takes in the number of training iterations for the generator rather than specifying the number of epochs for a dataset.

\begin{table*}[h]
\centering
\begin{tabular}{@{}cccccccc@{}}
\toprule
\textbf{Model} 
& \textbf{FID} 
& \textbf{Reported FID} 
& \textbf{$\mathbf{n_{\text{real}}}$}
& \textbf{$\mathbf{n_{\text{fake}}}$}
\\ \midrule
DCGAN & $28.95 \pm 0.42$ & $28.12^{\dagger} \cite{kurach2018large}$ & 10K & 10K \\
WGAN-GP & $26.08 \pm 0.12$ & $29.3 \cite{heusel2017gans}$ & 50K & 50K \\
SNGAN & $23.90 \pm 0.20$ & $21.7 \pm 0.21$ \cite{miyato2018spectral} & 10K & 5K \\
cGAN-PD & $17.84 \pm 0.17$ & $17.5$ \cite{miyato2018cgans} & 10K & 5K \\
SSGAN & $17.61 \pm 0.14$ & $17.88 \pm 0.64$ \cite{chen2018self} & 10K & 10K \\
InfoMax-GAN & $17.14 \pm 0.20$ & $17.39 \pm 0.08$ \cite{lee2019infomax} & 50K & 10K \\ \bottomrule
\end{tabular}
\vspace{0.1cm}
\caption{Mean CIFAR-10 FID scores of all implemented GANs against scores reported in the literature, as computed using the same sample sizes across 3 random seeds. To reproduce these scores we use the exact network architecture, training conditions, and evaluation methods in the respective papers. $\mathbf{n_{\text{real}}}$ and $\mathbf{n_{\text{fake}}}$ refer to the number of real and generated images used for computing FID, following the respective papers. $(\dagger)$: Best reported FID, where 10K real images are from test dataset. 
}
\label{tab:repro_table}
\end{table*}

\paragraph{Logging and visualizations.} As the training pipeline consists of multiple objects with different outputs, we adopt a simple message passing API by defining a \texttt{MetricLog} object, which is initialized at the start of training and collects information progressively down the pipeline. The resulting object is then passed to a \texttt{Logger} object, which prints out training information to the shell output, as well as writing TensorBoard summaries for visualizing loss and probability curves of the model, which are useful for debugging GAN training. Furthermore, we produce TensorBoard visualizations for two types of images every certain number of steps: (1) random images for checking diversity, and (2) images generated from a fixed set of noise vectors for checking the quality of images generated progressively. In order to sync these logging functionalities, we follow TensorFlow's \cite{abadi2016tensorflow} approach by adopting a global step variable that is shared amongst multiple objects.

\paragraph{Checkpointing experiments.} A common challenge in running a large set of experiments is that it is easy to forget the exact settings for an experiment. While descriptive folder names could be a solution, they can get unnecessarily long when many hyperparameters are used. Thus, we allow the \texttt{Trainer} object to store a JSON file of key hyperparameters used for a given experiment, such as the batch size and learning rate of the optimizers. Moreover, under the common scenario where a researcher wants to resume training from a checkpoint last saved, the \texttt{Trainer} object would check against the previous set of hyperparameters used and raise an error if there are discrepancies found. We support automatic restoration of checkpoint from the latest training step, and check that both discriminator and generator are restored from the same global step. Finally, we support dynamic stopping of experiments when needed by checkpointing models upon a keyboard interrupt.

\paragraph{Evaluation.} In order to ensure \textit{backward compatibility} of the obtained scores, we adopt the original implementations of all metrics \cite{openai_IS, ttur_FID, mmd_KID}. This is crucial since metrics like IS can be very sensitive to differences in model weights \cite{barratt2018note}. As these metrics involve the use of the same TensorFlow \cite{abadi2016tensorflow} Inception model pre-trained on ImageNet, we build a common interface shared by all metric computation methods. A PyTorch interface is then layered on top of these 3 metric computation methods in order for a user to compute scores without having to interact with the TensorFlow backend, yet ensuring the produced scores are compatible with those in existing literature.

\paragraph{Extensibility.} We rigorously test all implemented models and functions by building over 100 tests for this lightweight library, achieving a test coverage of more than 90\%. For continuous integration, we adopt CircleCI and CodeCov to ensure the library functionalities are maintained even when it is extended in the future.

\section{Experiments}
\subsection{Reproducibility}
To assess the correctness of our implementations, we evaluate our trained GANs using FID due to its popular usage in different papers, and compute scores using 3 random seeds to produce the mean and standard deviation. For all models, we train them using the same backbone architectures, training conditions, and evaluation method, as was done in the respective papers.  In Table \ref{tab:repro_table}, we show that all implementations can successfully reproduce the scores of the different papers. For WGAN-GP, our implementation performs better than the reported score, which uses the exact same ResNet backbone and training method as ours, and we find our FID can improve even further to $23.13 \pm 0.13$ when trained till 100K steps rather than 50K steps as was used in the reported results. Here, we emphasize that as FID produces highly biased estimates, it is important to reproduce the scores using the same number of samples to verify correctness.

\begin{table*}[ht]
\begin{tabular}{c|*{7}{c}}
\toprule
\multirow{2}{*}{\textbf{Metric}} 
& \multirow{2}{*}{\textbf{Dataset}} 
& \multirow{2}{*}{\textbf{Resolution}} 
& \multicolumn{4}{c}{\textbf{Models}} 
\\ \cmidrule(l){4-7} 
&  &  
& \textbf{SNGAN} 
& \textbf{cGAN-PD} 
& \textbf{SSGAN} 
& \textbf{InfoMax-GAN} 
\\ \midrule
\multirow{7}{*}{FID} 
 & LSUN-Bedroom & $128 \times 128$ & $25.87 \pm 0.03$ & - & $12.02 \pm 0.07$ & $12.13 \pm 0.16$ \\
 & CelebA & $128 \times 128$ & $12.93 \pm 0.04$ & - & $15.18 \pm 0.10$ & $9.50 \pm 0.04$ \\
 & CelebA & $64 \times 64$ & $5.71 \pm 0.02$ & - & $6.03 \pm 0.04$ & $5.71 \pm 0.06$ \\
 & STL-10 & $48 \times 48$ & $39.56 \pm 0.10$ & - & $37.06 \pm 0.19$ & $35.52 \pm 0.10$ \\
 & CIFAR-10 & $32 \times 32$ & $16.77 \pm 0.04$ & $10.84 \pm 0.03$ & $14.65 \pm 0.04$ & $15.12 \pm 0.10$ \\
 & CIFAR-100 & $32 \times 32$ & $22.61 \pm 0.06$ & $14.16 \pm 0.01$ & $22.18 \pm 0.10$ & $18.94 \pm 0.13$ \\
 & ImageNet & $32 \times 32$ & $23.04 \pm 0.06$ & $21.17 \pm 0.05$ & $21.79 \pm 0.09$ & $20.68 \pm 0.02$ \\
\midrule
\multirow{7}{*}{KID} 
 & LSUN-Bedroom & $128 \times 128$ & $0.0141 \pm 0.0001$ & - & $0.0077 \pm 0.0001$ & $0.0080 \pm 0.0001$ \\
 & CelebA & $128 \times 128$ & $0.0076 \pm 0.0001$ & - & $0.0101 \pm 0.0001$ & $0.0063 \pm 0.0001$ \\
 & CelebA & $64 \times 64$ & $0.0033 \pm 0.0001$ & - & $0.0036 \pm 0.0001$ & $0.0033 \pm 0.0001$ \\
 & STL-10 & $48 \times 48$ & $0.0369 \pm 0.0002$ & - & $0.0332 \pm 0.0004$ & $0.0326 \pm 0.0002$ \\
 & CIFAR-10 & $32 \times 32$ & $0.0125 \pm 0.0001$ & $0.0070 \pm 0.0001$ & $0.0101 \pm 0.0002$ & $0.0112 \pm 0.0001$ \\
 & CIFAR-100 & $32 \times 32$ & $0.0156 \pm 0.0003$ & $0.0085 \pm 0.0002$ & $0.0161 \pm 0.0002$ & $0.0135 \pm 0.0004$ \\
 & ImageNet & $32 \times 32$ & $0.0157 \pm 0.0002$ & $0.0145 \pm 0.0002$ & $0.0152 \pm 0.0002$ & $0.0149 \pm 0.0001$ \\
\midrule
\multirow{7}{*}{IS} 
& LSUN-Bedroom & $128 \times 128$ & $2.30 \pm 0.01$ & - & $2.12 \pm 0.01$ & $2.22 \pm 0.01$ \\
 & CelebA & $128 \times 128$ & $2.72 \pm 0.01$ & - & $2.63 \pm 0.01$ & $2.84 \pm 0.01$ \\
 & CelebA & $64 \times 64$ & $2.68 \pm 0.01$ & - & $2.67 \pm 0.01$ & $2.68 \pm 0.01$ \\
 & STL-10 & $48 \times 48$ & $8.04 \pm 0.07$ & - & $8.25 \pm 0.06$ & $8.54 \pm 0.12$ \\
 & CIFAR-10 & $32 \times 32$ & $7.97 \pm 0.06$ & $8.25 \pm 0.13$ & $8.17 \pm 0.06$ & $8.08 \pm 0.08$ \\
 & CIFAR-100 & $32 \times 32$ & $7.57 \pm 0.11$ & $8.92 \pm 0.07$ & $7.56 \pm 0.07$ & $7.86 \pm 0.10$ \\
 & ImageNet & $32 \times 32$ & $8.97 \pm 0.12$ & $9.08 \pm 0.17$ & $9.11 \pm 0.12$ & $9.04 \pm 0.10$ \\
\bottomrule
\end{tabular}
\vspace{0.1cm}
\caption{Baseline scores for different GANs across 7 datasets and 3 metrics. For cGAN-PD, we report scores only for datasets where labels are available. For FID and KID, lower is better. For IS, higher is better. We note that as LSUN-Bedroom and CelebA have only one class, the inception score is expected to be low.}
\label{tab:baselines}
\end{table*}

\begin{table*}[ht]
\vspace{.6cm}
\centering
\begin{tabular}{@{}ccccccccccc@{}}
\toprule
\textbf{Dataset} 
& \textbf{Resolution}
& \textbf{Split}
& \textbf{$\alpha$} 
& \textbf{$\beta_1$} 
& \textbf{$\beta_2$} 
& \textbf{Decay Policy} 
& $\mathbf{n_\text{dis}}$ 
& $\mathbf{n_\text{iter}}$ \\ \midrule
LSUN-Bedroom & $128 \times 128$ & Train & $2 \times 10^{-4}$ & $0.0$ & $0.9$ & None & $2$ & 100K \\
CelebA & $128 \times 128$ & Full & $2 \times 10^{-4}$ & $0.0$ & $0.9$ & None & $2$ & 100K \\
CelebA & $64 \times 64$ & Full & $2 \times 10^{-4}$ & $0.0$ & $0.9$ & Linear & $5$ & 100K \\
STL-10 & $48 \times 48$ & Unlabeled & $2 \times 10^{-4}$ & $0.0$ & $0.9$ & Linear & $5$ & 100K \\
CIFAR-10 & $32 \times 32$ & Train & $2 \times 10^{-4}$ & $0.0$ & $0.9$ & Linear & $5$ & 100K \\
CIFAR-100 & $32 \times 32$ & Train & $2 \times 10^{-4}$ & $0.0$ & $0.9$ & Linear & $5$ & 100K \\
ImageNet & $32 \times 32$ & Train & $2 \times 10^{-4}$ & $0.0$ & $0.9$ & Linear & $5$ & 100K \\ \bottomrule
\end{tabular}
\vspace{0.1cm}
\caption{Training configurations for each dataset. Split refers to the dataset split selected for training; $\alpha$ refers to the learning rate, and $(\beta_1, \beta_2)$ are the hyperparameters of the Adam \cite{kingma2014adam} optimizer; decay policy refers to the learning rate scheduling; $\mathbf{n_\text{dis}}$ represents the number of discriminator steps taken per generator step, and $\mathbf{n_\text{iter}}$ refers to the number of training iterations.}
\label{tab:training_config}
% \vspace{-5cm}
\end{table*}

\begin{table*}[!h]
\vspace{.5cm}
\centering
\begin{tabular}{@{}cccc@{}}
\toprule
\textbf{Metric} & $\mathbf{n_\text{real}}$ & $\mathbf{n_\text{fake}}$ & \textbf{Remarks} \\ 
\midrule
Inception Score (IS) & - & 50K & 10 splits. \\
Fréchet Inception Distance (FID) & 50K & 50K & - \\
Kernel Inception Distance (KID) & 50K & 50K & 10 splits. \\
\bottomrule
\end{tabular}
\vspace{0.1cm}
\caption{Evaluation method for each metric used to compute the baseline scores. $\mathbf{n_{\text{real}}}$ and $\mathbf{n_{\text{fake}}}$ refer to the number of real and generated images, respectively.}
\label{tab:eval_config}
\end{table*}

\subsection{Baselines}
For all implementations, we adopt a Residual Network \cite{he2016deep} backbone following \cite{miyato2018cgans, miyato2018spectral, chen2018self, gulrajani2017improved,lee2019infomax} that has been shown to consistently work well in practice. Due to computational constraints, we provide only baseline results for SNGAN, cGAN-PD, SSGAN, and InfoMax-GAN, but with plans to expand our results in the future. For each dataset, we train all GANs using the same training configurations as detailed in Table \ref{tab:training_config}, and evaluate them using the same method as detailed in Table \ref{tab:eval_config}. In Table \ref{tab:baselines} we provide the mean baseline scores of the GANs on different datasets and metrics, as computed across 3 random seeds. In general, we observe that the baseline performances reflect the expected relative performances of different GANs. As cGAN-PD is a conditional GAN with the benefit of having labeled information, it performs generally the best across all datasets. SSGAN and InfoMax-GAN builds upon SNGAN, but with different auxiliary tasks for improving GAN training, resulting in better performance over the SNGAN baseline. Finally, while SSGAN and InfoMax-GAN performs similarly for most datasets, we observe InfoMax-GAN is able to achieve significant improvements in datasets where SSGAN did not, such as in the high resolution CelebA dataset where SSGAN did not perform better than the SNGAN baseline, agreeing with the reported phenomenon in \cite{chen2018self}.

Through providing these scores and the exact method to obtain them transparently, we aim to reduce the need for researchers to cross-cite many different papers with varying implementations in order to compare scores. Furthermore, these baseline results are useful from a practitioner's point of view: given a fixed computational budget, it is possible to decide which GAN can perform the best. While we have only included one set of hyperparameters per dataset, we aim to include more settings as future work. To aid reproducibility, we include a model zoo with pre-trained checkpoints to accompany these results.
\vspace{-0.1cm}
\section{Discussion}
In this work, we have introduced Mimicry, a PyTorch library targeted at the reproducibility of GAN research through providing comprehensively evaluated baseline performances and standardized tools for researchers to build upon. We hope through producing clear and fair comparisons with past works, GAN research findings can be made more reproducible. As future work, we aim to cover a wider range of GANs and metrics \cite{kynkaanniemi2019improved, ravuri2019classification, grnarova2019domain}, and expand to include other GAN tasks, such as 3D view synthesis. Finally, the library is released as open source software under the MIT License for use by researchers.

% \clearpage
{\small
\bibliographystyle{ieee_fullname}
\bibliography{references}
}

\end{document}